\documentclass{article}
\usepackage{spconf,amsmath,graphicx}
\usepackage{url}
\usepackage{amssymb}
\usepackage{epsfig}
\usepackage{balance}
\usepackage{booktabs}
\usepackage{cite}
\usepackage{multirow}
\usepackage[misc]{ifsym}
\usepackage{bm}

\title{Adversarial Contrastive Distillation with Adaptive Denoising}
\name{Yuzheng Wang, Zhaoyu Chen, Dingkang Yang, Yang Liu , Siao Liu, Wenqiang Zhang$^{\textrm{\Letter}}$, Lizhe Qi$^{\textrm{\Letter}}$\thanks{\textrm{\Letter}\;\;The corresponding authors are Lizhe Qi and Wenqiang Zhang. This work is supported by Natural Science Foundation of Jiangxi Province (No.20212BAB202026), Shanghai Municipal Science and Technology Major Project (No.2021SHZDZX0103), the Shanghai Engineering Research Center of AI \& Robotics, Fudan University, China, and the Engineering Research Center of AI \& Robotics, Ministry of Education, China.}}

\address{Academy for Engineering \& Technology, Fudan University, Shanghai, China}

\begin{document}

\maketitle

\begin{abstract}
Adversarial Robustness Distillation (ARD) is a novel method to boost the robustness of small models. 
Unlike general adversarial training, its robust knowledge transfer can be less easily restricted by the model capacity.
However, the teacher model that provides the robustness of knowledge does not always make correct predictions, interfering with the student's robust performances.
Besides, in the previous ARD methods, the robustness comes entirely from one-to-one imitation, ignoring the relationship between examples.
To this end, we propose a novel structured ARD method called Contrastive Relationship DeNoise Distillation (CRDND).
We design an adaptive compensation module to model the instability of the teacher.
Moreover, we utilize the contrastive relationship to explore implicit robustness knowledge among multiple examples.
Experimental results on multiple attack benchmarks show CRDND can transfer robust knowledge efficiently and achieves state-of-the-art performances.

\end{abstract}


\section{Introduction}

Deep learning models have achieved great success in computer vision \cite{yang2023target,liu2022learninga,Chen_2022_CVPR,9987686,wang2023explicit}, signal processing \cite{yang2022disentangled,yang2022learning,liu2022appearance}, and other fields \cite{huang2022cmua}.
These models, however, can usually be attacked by adding small permutations to natural inputs \cite{liu2022efficient,chen2022shape,wang2022boosting}.
The vulnerability has aroused people's concern about applying deep learning technology in automatic driving, financial forecasting, and face fingerprint detection. 
Concurrently, this also helps researchers to rethink the robustness of the model \cite{yan2020hijacking}.

Recently, many defense strategies have emerged to improve the adversarial robustness of models, such as data processing and model training methods.
Among them, Adversarial Training (AT) is recognized as the most effective defensive strategy \cite{madry2017towards}.
AT takes the adversarial examples as a kind of data enhancement so that the model can learn the defense strategy against the potential attack threat.
Despite the outstanding performance, AT continuously expands the training dataset resulting in expensive model training costs.
In addition, the effectiveness of AT is often related to the model capacity \cite{zi2021revisiting}. 
The robustness of small models is often limited, which makes it challenging to apply this technology to micro-robots, mobile phones, and driverless cars.
All these have led to introducing knowledge distillation to improve AT, called Adversarial Robustness Distillation (ARD).

Goldblum \emph{et al}.\cite{goldblum2020adversarially} propose the concept of ARD. 
They show that the robust model can avoid the expensive AT cost. 
On the contrary, by transferring the knowledge of the pre-trained robust model, the small model can obtain a higher robustness performance than the standard robust training.
Zhu \emph{et al}.\cite{zhu2022reliable} propose a multi-stage strategy to improve the efficiency of knowledge transfer further, thus improving the robustness of the student.
Zi \emph{et al}.\cite{zi2021revisiting} think the soft target label is essential in robustness distillation, so they use the entirely soft target label of a large robustness model to replace one-hot label to help the student further improve robustness.

Although the previous ARD methods can avoid the expensive AT cost, there are still many issues.
On the one hand, the teacher's predictions are not always correct. 
Especially with the student's progress, the confidence level of the teacher's predictions for the adversarial examples generated by the student will gradually decrease \cite{zhu2022reliable}.
As the main source of robust knowledge, this unstable prediction limits the student's performance.
Zhu \emph{et al}.\cite{zhu2022reliable} try to model this instability, but in fact, this instability is closely related to the capacity of the backbones used by the teacher and student.
Hence, their methods are not universal for all backbone pairs.
On the other hand, all previous methods can be summarized as one-to-one naive example imitation learning \cite{park2019relational}.
Therefore, the robust knowledge comes entirely from the pre-trained teacher, which ignores the implicit similarity between multiple examples.
Moreover, the learning potential of the student model may not be fully developed, which is shown by the fact that the teacher's performance limits the student's robust accuracy.

\begin{figure*}[t]
	\centering
	\includegraphics[scale=0.285]{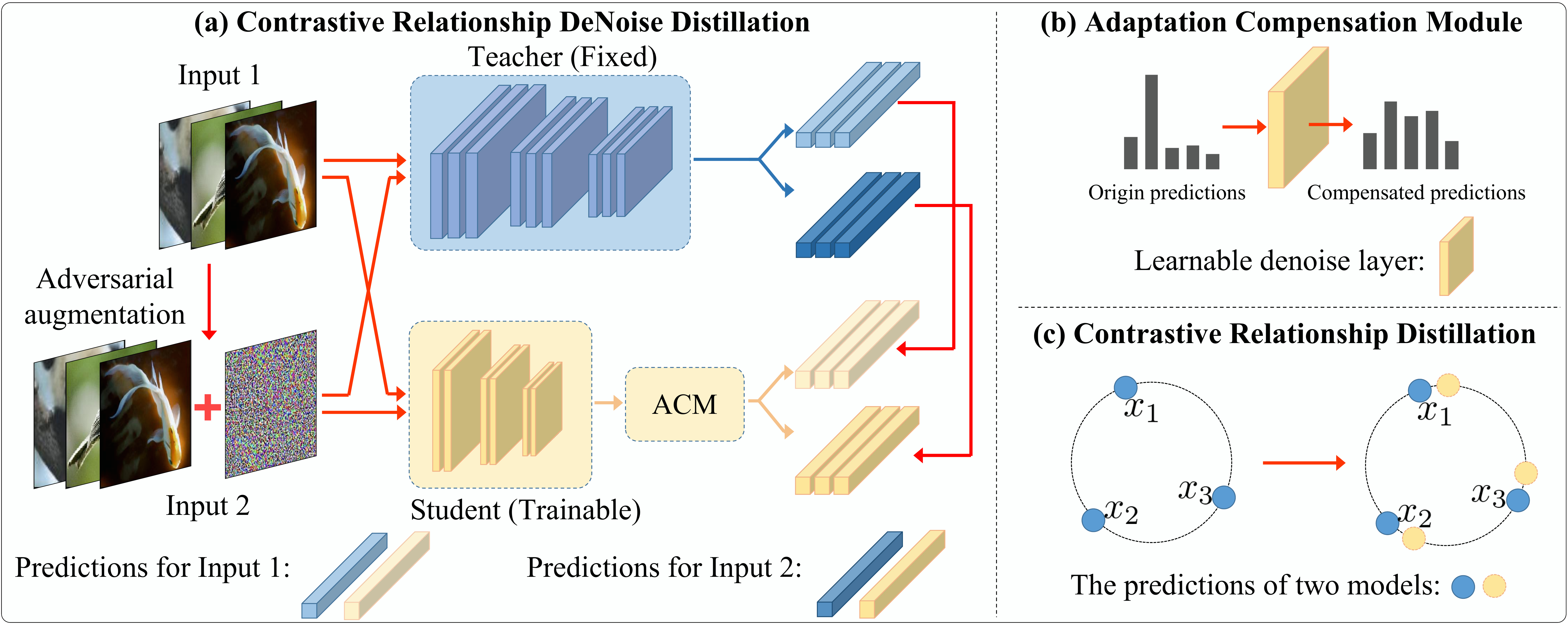}
	\vspace{-0.1cm}
	\caption{(a) Overview of our Contrastive Relationship DeNoise Distillation. (b) The proposed adaptive compensation module (ACM). The columns denote the predictions. (c) The proposed contrastive relationship distillation method.}
	\label{fig1}
	\vspace{-0.3cm}
\end{figure*}

In this paper, we propose a novel structured adversarial robustness distillation method called Contrastive Relationship DeNoise Distillation (CRDND).
Specifically, considering the unstable teacher's predictions, we design an adaptive compensation module to help the student correct possible prediction noise through the learnable robustness layer.
We then introduce the idea of contrastive learning into the field of ARD to model the robustness relationship among multiple examples.
The student model can simultaneously learn the knowledge of the robust teacher and sufficiently explore the knowledge from examples to improve performance.
The main contributions of this work are summarized as follows:
\vspace{-0.1cm}
\begin{itemize}
\item We propose a novel adversarial robustness distillation method called Contrastive Relationship DeNoise Distillation. 
The structured relationship among multiple examples replaces one-to-one imitation learning to help the student achieve better results.
\vspace{-0.1cm}
\item To restrain the influence of the teacher's unstable prediction, we design a plug-and-play adaptive compensation module.
The possible prediction noise of the teacher is refined through the learnable denoise layer.
\vspace{-0.5cm}
\item Experimental results against multiple adversarial attacks show that our CRDND method achieves state-of-the-art performances.
As a result, the robustness of the small model is greatly improved.

\end{itemize}

\section{METHODOLOGY}

\subsection{Overview}

The overview of the proposed Contrastive Relationship DeNoise Distillation is shown in Fig.\;\ref{fig1}(a).
Following the previous ARD methods, we assume that natural examples and the fixed and pre-trained robust teacher model (e.g., WideResNet \cite{zagoruyko2016wide}) are available.
Then, the goal is to train a small student model (e.g., MobileNetV2 \cite{sandler2018mobilenetv2}) while inheriting the robustness of the teacher.
The input includes natural and adversarial examples to help the student deal with multiple examples scenarios.
To overcome the uncertainty of the teacher's prediction, we design an Adaptive Compensation Module (ACM) to model the instability. 
A learnable denoise layer after the logit predictions of the student model is added to estimate the correctness of the teacher's answers.
To improve the efficiency of knowledge transfer, we denote the Contrastive Relationship Distillation of natural and adversarial examples, respectively, to deeply explore the knowledge among multiple examples. 
As a result, our method is not completely limited by the robustness of the teacher and achieves good performance.

\vspace{-0.2cm}
\subsection{Adaptive Compensation Module}

Similar to traditional knowledge distillation \cite{hinton2015distilling}, we transfer knowledge by constraining the teacher's and student's predictions. 
We denote $f_{T}$ and $ f_{S}$ as the logits predictions of the teacher and student models, $x$ as the natural examples, and ${x}'$ as the adversarial examples.
The process of knowledge transfer can be expressed as:
\vspace{-0.05cm}
\begin{equation}\label{eq1}
\mathcal{L}=\sum_{x,x' \in \mathcal{X}} D(f_{T}(x,{x}'), f_{S}(x,{x}')),
\end{equation}
where $D$ is a distance representation.

However, as mentioned above, the prediction of the teacher model is not necessarily correct. 
Incorrect predictions often lead to incorrectly information for the student to learn \cite{zhu2022reliable}.
Therefore, we define a learnable noise layer $M\in \mathbb{R}^{k\times k}$ to model the instability of the teacher's prediction as shown in Fig.\;\ref{fig1}(b).
It represents the correct probabilities of the teacher's predictions.
We regulate parameters in $M$ with the estimate of the teacher's accuracy.
Specifically, we calculate the accuracy of the teacher model in the current training epoch about natural or adversarial examples set.
The setting rule of M is: the main class weight is represented by the current accuracy rate, and the rest of the class is averaged (the sum is 1).
The calculated value is used as an estimate of the teacher's true accuracy in the current training epoch.
The column of $M$ can be regarded as a probability distribution, satisfying $ {\textstyle \sum_{j=1}^{k}} M_{ij}=1$, where $k$ is the number of classes.
$M$ is denoted as:
$$
M=\begin{bmatrix}
 a_1 & \frac{1-a_2}{k-1}   & \cdots  & \frac{1-a_k}{k-1} \\
 \frac{1-a_1}{k-1} & a_2 & \cdots  &\frac{1-a_k}{k-1} \\
 \vdots  & \vdots &  \ddots & \vdots \\
 \frac{1-a_1}{k-1} & \frac{1-a_2}{k-1} &  \cdots & a_k
\end{bmatrix}, \eqno{(2)}
$$where $a_i$ denotes the accuracy of $i$-th class. 
$M_1,M_2$ denote the noise layers for natural and adversarial scenarios.
Then Eq.\;\ref{eq1} can be rewritten as:
$$ \mathcal{L'}\!=\!\sum_{x,x' \in \mathcal{X}} \!\! D(f_{T}(x,{x}'), M_1/M_2(f_{S}(x,{x}'))). \eqno{(3)} $$Since our method only needs to estimate the teacher's current accuracy, it applies to various teacher-student backbone combinations.
By modeling the instability of the teacher, our method can also overcome the dilemma that the reliability of the teacher declines gradually during the training epoch.

\subsection{Contrastive Relationship Distillation}

Although Eq.\;3 can compensate the mistakes that the teacher may make when transferring robustness knowledge, such knowledge depends entirely on the teacher.
To further explore the structured knowledge among multiple examples, we propose a contrastive relationship distillation method to replace the above process as shown in Fig.\;\ref{fig1}(c). 
Specifically, we focus on the consistency of the teacher's and student's example predictions in a mini-batch.
Concurrently, we hope to separate two predictions that are not corresponding.
Based on these, we build two kinds of structured knowledge to respond to natural and adversarial scenarios.

Firstly, for natural examples $\bm{x}=\left \{ x_1,\dots,x_n \right \}$, we first obtain predictions from robust teacher $f_T(x_i)$ and noise compensated student $M_1(f_S(x_i))$.
Then, the contrastive relationship can be represented as:
$$ \ell_{nat}^{x_{i}}=\frac{\exp( \cos (M_1(f_{S}(x_{i})), f_{T}(x_{i}))  / \tau_{1} )}{ {\textstyle \sum_{k=1,k\ne i}^{2N}} \exp( \cos (M_1(f_{S}(x_{k})), f_{T}(x_{i}))/ \tau_{1}) }, \eqno{(4)} $$where $\tau_{1}$ denotes temperature parameter and $N$ denotes the batch size.
Next, we can calculate the relationship distillation loss for natural examples as:
$$ \mathcal{L}_{nat}=-\frac{1}{N}\sum_{j=1}^{N}  \log_{}{\ell_{nat}^{x_{j}}}. \eqno{(5)}$$

For adversarial examples $\bm{x'}$, the knowledge representation and transfer process are similar to the above:
$$ \ell_{adv}^{x'_{i}}=\frac{\exp( \cos (M_2(f_{S}(x'_{i})), f_{T}(x'_{i}))  / \tau_{2} )}{ {\textstyle \sum_{k=1,k\ne i}^{2N}} \exp( \cos (M_2(f_{S}(x'_{k})), f_{T}(x'_{i}))/ \tau_{2}) }, \eqno{(6)} $$
$$ \mathcal{L}_{adv}=-\frac{1}{N}\sum_{j=1}^{N}  \log_{}{\ell_{adv}^{x'_{j}}}.  \eqno{(7)}$$

Finally, we can get the total Contrastive Relationship DeNoise Distillation loss as:
$$
\mathcal{L}_{total}=\lambda\! \cdot \!\mathcal{L}_{nat}+(1-\lambda)\! \cdot \! \mathcal{L}_{adv},   \eqno{(8)}
$$where the $\lambda$ is the loss trade-off parameter. Unlike Eq.\;3, Eq.\;8 models the consistency between the teacher and student and the difference among multiple examples.
The knowledge from multiple examples is crucial, especially when the teacher model cannot give reliable predictions.

Especially, it is worth noting that our contrastive learning method differs from previous methods.
First, our method does not rely on large negative examples sets, such as a large memory bank \cite{wu2018unsupervised}, large divisions \cite{he2020momentum}, or a large batch size \cite{chen2020simclr}.
It also does not rely on additional normalization and pre-training network update \cite{caron2021emerging} with high computing costs.
Second, our method does not need to design suitable data augmentation operators \cite{chen2020simclr} carefully.
As a result, our method is simple and computationally efficient.

\section{EXPERIMENTS}

\subsection{Experimental Settings}

We evaluate proposed CRDND method on CIFAR-10, and CIFAR-100 \cite{krizhevsky2009learning}, the commonly used adversarial robustness test datasets.
The baseline methods consider two AT methods: SAT \cite{madry2017towards}, TRADES \cite{zhang2019theoretically}, three ARD methods: ARD \cite{goldblum2020adversarially}, IAD \cite{zhu2022reliable}, RSLAD \cite{zi2021revisiting}, and a natural training method.

\noindent\textbf{Teacher and Student.} For fair comparison, we choose the same teacher models following RSLAD \cite{zi2021revisiting} including WideResNet-34-10 \cite{zagoruyko2016wide} for CIFAR-10 and WideResNet-70-16 \cite{gowal2020uncovering} for CIFAR-100. 
The teacher model is fixed during the whole training process.
Besides, we set two backbones of the students including ResNet-18 \cite{he2016deep} and MobileNetV2 \cite{sandler2018mobilenetv2}.

\noindent\textbf{Implementation Details.} The proposed model is implemented in PyTorch and trained on eight RTX 2080 Ti GPUs.
We set the loss trade-off parameter $\lambda$ as 0.2, and the temperature parameters $\tau_{1}, \tau_{2}$ as 0.5.
The student is trained via SGD optimizer with cosine annealing learning rate with an initial value of 0.1, momentum 0.9 and weight decay 2e-4. 
The batch size is 128, and the total number of training epochs is 300, the same as the previous works.
For other baseline methods, we follow the setting of RSLAD \cite{zi2021revisiting}.

\noindent\textbf{Attacks Evaluation.}
We evaluate the model against multiple adversarial attacks: FGSM \cite{goodfellow2014explaining}, PGD${\rm _{SAT}}$ (PGD$\rm _{S}$) \cite{madry2017towards}, PGD${\rm _{TRADES}}$ (PGD$\rm _{T}$) \cite{zhang2019theoretically} and AutoAttack (AA) \cite{croce2020reliable}.
Besides, the above attack methods are the same as the settings of RSLAD \cite{zi2021revisiting}.

\begin{table*}[t]
\centering
\caption{Adversarial robustness accuracy (\%) on CIFAR-10 and CIFAR-100 datasets. The maximum adversarial perturbation $\epsilon$ is 8/255. RN-18 and MN-V2 are abbreviations of ResNet-18 and MobileNetV2 respectively.
\textbf{Bold} and \underline{underline} numbers denote the best and the second best results, respectively.}
\label{tab1}
\setlength{\tabcolsep}{2.9mm}
\scalebox{0.82}{
\begin{tabular}{@{}ccccccc|ccccccc@{}}
\toprule[1pt]
\multicolumn{7}{c|}{CIFAR-10} &
  \multicolumn{7}{c}{CIFAR-100} \\ \midrule
\multicolumn{1}{c|}{\multirow{2}{*}{Model}} &
  \multicolumn{1}{c|}{\multirow{2}{*}{Method}} &
  \multicolumn{5}{c|}{Attacks Evaluation} &
  \multicolumn{1}{c|}{\multirow{2}{*}{Model}} &
  \multicolumn{1}{c|}{\multirow{2}{*}{Method}} &
  \multicolumn{5}{c}{Attacks Evaluation} \\
\multicolumn{1}{c|}{} &
  \multicolumn{1}{c|}{} &
  Clean &
  FGSM &
  PGD${\rm _{S}}$ &
  PGD${\rm _{T}}$ &
  AA &
  \multicolumn{1}{c|}{} &
  \multicolumn{1}{c|}{} &
  Clean &
  FGSM &
  PGD${\rm _{S}}$ &
  PGD${\rm _{T}}$ &
  AA \\ \midrule
\multicolumn{1}{c|}{\multirow{7}{*}{RN-18}} &
  \multicolumn{1}{c|}{Nature} &
  \textbf{94.65} &
  19.26 &
  0.0 &
  0.0 &
  0.0 &
  \multicolumn{1}{c|}{\multirow{7}{*}{RN-18}} &
  \multicolumn{1}{c|}{Nature} &
  \textbf{75.55} &
  9.48 &
  0.0 &
  0.0 &
  0.0 \\
\multicolumn{1}{c|}{} &
  \multicolumn{1}{c|}{SAT} &
  83.38 &
  56.41 &
  49.11 &
  51.11 &
  45.83 &
  \multicolumn{1}{c|}{} &
  \multicolumn{1}{c|}{SAT} &
  57.46 &
  28.56 &
  24.07 &
  25.39 &
  21.79 \\
\multicolumn{1}{c|}{} &
  \multicolumn{1}{c|}{TRADES} &
  81.93 &
  57.49 &
  52.66 &
  53.68 &
  49.23 &
  \multicolumn{1}{c|}{} &
  \multicolumn{1}{c|}{TRADES} &
  55.23 &
  30.48 &
  27.79 &
  28.53 &
  23.94 \\
\multicolumn{1}{c|}{} &
  \multicolumn{1}{c|}{ARD} &
  83.93 &
  59.31 &
  52.05 &
  54.20 &
  49.19 &
  \multicolumn{1}{c|}{} &
  \multicolumn{1}{c|}{ARD} &
  \underline{60.64} &
  33.41 &
  29.16 &
  30.30 &
  25.65 \\
\multicolumn{1}{c|}{} &
  \multicolumn{1}{c|}{IAD} &
  83.24 &
  58.60 &
  52.21 &
  54.18 &
  49.10 &
  \multicolumn{1}{c|}{} &
  \multicolumn{1}{c|}{IAD} &
  57.66 &
  33.26 &
  29.59 &
  30.58 &
  25.12 \\
\multicolumn{1}{c|}{} &
  \multicolumn{1}{c|}{RSLAD} &
  83.38 &
  \underline{60.01} &
  \underline{54.24} &
  \underline{55.94} &
  \textbf{51.49} &
  \multicolumn{1}{c|}{} &
  \multicolumn{1}{c|}{RSLAD} &
  57.74 &
  \underline{34.20} &
  \underline{31.08} &
  \underline{31.90} &
  \underline{26.70} \\
\multicolumn{1}{c|}{} &
  \multicolumn{1}{c|}{\textbf{CRDND}} &
  \underline{84.11} &
  \textbf{64.24} &
  \textbf{59.91} &
  \textbf{61.25} &
  \underline{49.88} &
  \multicolumn{1}{c|}{} &
  \multicolumn{1}{c|}{\textbf{CRDND}} &
  59.00 &
  \textbf{38.02} &
  \textbf{35.29} &
  \textbf{36.29} &
  \textbf{27.05} \\ \midrule[1pt]
\multicolumn{1}{c|}{\multirow{7}{*}{MN-V2}} &
  \multicolumn{1}{c|}{Nature} &
  \textbf{92.95} &
  14.47 &
  0.0 &
  0.0 &
  0.0 &
  \multicolumn{1}{c|}{\multirow{7}{*}{MN-V2}} &
  \multicolumn{1}{c|}{Nature} &
  \textbf{74.58} &
  7.19 &
  0.0 &
  0.0 &
  0.0 \\
\multicolumn{1}{c|}{} &
  \multicolumn{1}{c|}{SAT} &
  82.48 &
  56.44 &
  50.10 &
  51.74 &
  46.32 &
  \multicolumn{1}{c|}{} &
  \multicolumn{1}{c|}{SAT} &
  56.85 &
  31.95 &
  28.33 &
  29.50 &
  24.71 \\
\multicolumn{1}{c|}{} &
  \multicolumn{1}{c|}{TRADES} &
  80.57 &
  56.05 &
  51.06 &
  52.36 &
  47.17 &
  \multicolumn{1}{c|}{} &
  \multicolumn{1}{c|}{TRADES} &
  56.20 &
  31.37 &
  29.21 &
  29.83 &
  24.16 \\
\multicolumn{1}{c|}{} &
  \multicolumn{1}{c|}{ARD} &
  83.20 &
  58.06 &
  50.86 &
  52.87 &
  48.34 &
  \multicolumn{1}{c|}{} &
  \multicolumn{1}{c|}{ARD} &
  \underline{59.83} &
  33.05 &
  29.13 &
  30.26 &
  25.53 \\
\multicolumn{1}{c|}{} &
  \multicolumn{1}{c|}{IAD} &
  81.91 &
  57.00 &
  51.88 &
  53.23 &
  48.40 &
  \multicolumn{1}{c|}{} &
  \multicolumn{1}{c|}{IAD} &
  56.14 &
  32.81 &
  29.81 &
  30.73 &
  25.74 \\
\multicolumn{1}{c|}{} &
  \multicolumn{1}{c|}{RSLAD} &
  83.40 &
  \underline{59.06} &
  \underline{53.16} &
  \underline{54.78} &
  \textbf{50.17} &
  \multicolumn{1}{c|}{} &
  \multicolumn{1}{c|}{RSLAD} &
  58.97 &
  \underline{34.03} &
  \underline{30.40} &
  \underline{31.36} &
  \underline{26.12} \\
\multicolumn{1}{c|}{} &
  \multicolumn{1}{c|}{\textbf{CRDND}} &
  \underline{83.89} &
  \textbf{65.25} &
  \textbf{59.93} &
  \textbf{61.33} &
  \underline{48.79} &
  \multicolumn{1}{c|}{} &
  \multicolumn{1}{c|}{\textbf{CRDND}} &
  58.60 &
  \textbf{38.03} &
  \textbf{36.05} &
  \textbf{37.02} &
  \textbf{26.56} \\ \bottomrule[1pt]
\end{tabular}
}
\vspace{-0.35cm}
\end{table*}

\vspace{-0.1cm}
\subsection{Comparison to State-of-the-art Methods}

The robustness performances of our and other baseline methods are shown in Table \ref{tab1}.
We compare the best checkpoint of various methods.
The `Nature' training method is selected based on the performance of clean test examples.
Besides, other training methods are selected based on the robustness performance against PGD${\rm _{T}}$ following previous methods.
It can be seen from the results that our CRDND method achieves state-of-the-art robustness performances on multiple benchmarks.
Especially for FGSM and PGD evaluation metrics, our method has greatly improved (4\%-6\%).
In general, the performance of adversarial robustness is in a trade-off relationship with the performance of clean conditions unless ground truth labels are used (e.g., Nature and ARD methods).
Our method is particularly competitive in both conditions without any labels, which shows that our student improves overall performance by deeply exploring additional knowledge among multiple examples.

\begin{table}[h]
\vspace{-0.45cm}
\centering
\caption{Ablation studies on CIFAR-100 dataset (\%).}
\label{tab2}
\setlength{\tabcolsep}{2.9mm}
\scalebox{0.8}{
\begin{tabular}{c|c|c|cccc}
\hline
\multirow{2}{*}{ID} & \multirow{2}{*}{Model} & \multirow{2}{*}{Method} & \multicolumn{4}{c}{Attacks Evaluation}                              \\
                    &                        &                         & FGSM           & PGD${\rm _{S}}$ & PGD${\rm _{T}}$ & AA             \\ \hline
1                  & \multirow{2}{*}{RN-18} & \textbf{Ours}           & \textbf{38.02} & \textbf{35.29}  & \textbf{36.29}  & \textbf{27.05} \\
2                  &                        & w/o ACM                 & 38.02          & 35.14           & 36.11           & 26.29          \\ \cline{2-7} 
3                  & \multirow{2}{*}{MN-V2} & \textbf{Ours}           & \textbf{38.03} & \textbf{36.05}  & \textbf{37.02}  & \textbf{26.56} \\
4                  &                        & w/o ACM                 & 37.91          & 35.80           & 36.78           & 26.37          \\ \hline
5                  & \multirow{2}{*}{RN-18} & w/o $\mathcal{L}_{nat}$ & 37.41          & 34.87           & 35.84           & 25.20          \\
6                  &                        & w/o $\mathcal{L}_{adv}$ & 34.58          & 29.41           & 30.40           & 26.33          \\ \cline{2-7} 
7                  & \multirow{2}{*}{MN-V2} & w/o $\mathcal{L}_{nat}$ & 37.80          & 35.85           & 36.71           & 25.67          \\
8                  &                        & w/o $\mathcal{L}_{adv}$ & 34.97          & 29.56           & 30.78           & 26.14          \\ \hline
\end{tabular}
}
\vspace{-0.75cm}
\end{table}

\subsection{Ablation Study}

To verify the effectiveness of our proposed Adaptive Compensation Module (ACM), we set the baselines to bold (Ours: full CRDND) and discard ACM to demonstrate the impact on the results.
Table \ref{tab2} (1-4) contrasts the impacts of model robustness with or without ACM (the w/o in the table means without).
When the ACM is discarded, the robustness of students decreases to varying degrees on various attack metrics.
We believe that the decline is due to the lack of estimation of the prediction accuracy of the teacher model, which indicates that the frequent incorrect predictions given by the teacher model may interfere with the student's learning.

To verify the two optimization objectives we designed,  we separate them to test the effectiveness of using them separately.
Table \ref{tab2} (5-8) shows the robustness performance without $\mathcal{L}_{nat}$ or $\mathcal{L}_{adv}$. 
Compared with Table \ref{tab2} (1, 3), we observe that the student's performance will decline no matter which objective is missing.
It is worth noting that when only $\mathcal{L}_{nat}$ is used, the student model does not directly learn any knowledge about adversarial examples. 
We analyze that the robustness here comes from our structured relationship distillation method, which can transfer the robust information among examples by learning a relative relationship.

\section{CONCLUSION}

In this paper, we propose a novel adversarial robustness distillation method called Contrastive Relationship DeNoise Distillation (CRDND), which eliminates the previous methods' complete trust and dependence on the teacher model.
Firstly, we design a plug-and-play Adaptive Compensation Module, which corrects noise knowledge by estimating the possible prediction errors of the teacher network.
Secondly, we propose a novel idea to improve the models' robustness by exploring implicit knowledge among multiple examples to deal with adversarial attacks.
Experimental results on multiple attack benchmarks show CRDND not only significantly improves the robustness performance of the student model but also retains the performance of the clean examples.

\footnotesize 
{
\bibliographystyle{IEEEbib}
\bibliography{refs}
}

\end{document}